\documentclass{article}

\usepackage{PRIMEarxiv}

\usepackage[utf8]{inputenc} 
\usepackage[T1]{fontenc}    
\usepackage{hyperref}       
\usepackage{url}            
\usepackage{booktabs}       
\usepackage{amsfonts}       
\usepackage{nicefrac}       
\usepackage{microtype}      
\usepackage{lipsum}
\usepackage{fancyhdr}       
\usepackage{graphicx}       
\graphicspath{{media/}}     

\title{A New Benchmark for Evaluating Automatic Speech Recognition in the Arabic Call Domain}

\author{
  Qusai Abo Obaidah, Muhy Eddin Za'ter, Adnan Jaljuli, Ali Mahboub, Asma Hakouz, Bashar Al-Rfooh and Yazan Estaitia\\
  Maqsam \\
  Amman, Jordan\\
  \texttt{\{qusai,muhyeddin,adnan.jaljuli,ali.mahbob, asma, bashar.alrfooh, yazan.estaitia\}@maqsam.com} \\
}

\begin{document}
\maketitle

\begin{abstract}
This work is an attempt to introduce a comprehensive benchmark for Arabic speech recognition, specifically tailored to address the challenges of telephone conversations in Arabic language. Arabic, characterized by its rich dialectal diversity and phonetic complexity, presents a number of unique challenges for automatic speech recognition (ASR) systems. These challenges are further amplified in the domain of telephone calls, where audio quality, background noise, and conversational speech styles negatively affect recognition accuracy. Our work aims to establish a robust benchmark that not only encompasses the broad spectrum of Arabic dialects but also emulates the real-world conditions of call-based communications. By incorporating diverse dialectical expressions and accounting for the variable quality of call recordings, this benchmark seeks to provide a rigorous testing ground for the development and evaluation of ASR systems capable of navigating the complexities of Arabic speech in telephonic contexts. This work also attempts to establish a baseline performance evaluation using state-of-the-art ASR technologies.
\end{abstract}


\section{Introduction}

The emergence of automatic speech recognition (ASR) systems has revolutionized the way people interact with machines, enabling more natural and efficient communication. While ASR endeavors have made significant advancements, they continue to face considerable challenges when dealing with the Arabic language, especially in the context of telephone calls. Arabic's rich diversity of dialects, rich morphology and the inherent complexities of spoken communication within calls\cite{farghaly2009arabic}, such as colloquialisms, code-switching with English, and quality related issues such as varying speech rates, and background noise, pose unique obstacles to accurate speech recognition \cite{fernandez2022study}.

Therefore, the development of an effective benchmark for Arabic speech recognition in call domains is crucial for advancing ASR technologies. Such a benchmark must accurately reflect the linguistic diversity and the specific conditions encountered in telephone communications to ensure the development of robust and reliable ASR systems. This paper presents the creation of a benchmark collected from real-world data designed to meet these requirements, focusing on capturing the wide range of the above-mentioned aspects.

Furthermore, to effectively evaluate the performance of ASR systems against the challenges of Arabic speech calls, this study focuses on establishing the Word Error Rate (WER) and Character Error Rate (CER) for state-of-the-art models and commercial APIs using the developed benchmark.

The rest of the paper is divided as follows; section 2 presents a brief overview of the available literature, where section 3 describes the dataset collection and preparation pipeline. Section 4 presents the models evaluated. Finally section 5 presents the results of the ASR models on the developed benchmark.

\section{Literature Review}

Despite Arabic being one of the most widely spoken languages globally, with over 420 million speakers, the development of robust ASR systems tailored to its unique characteristics has lagged behind other languages, mainly due to its linguistic complexity and diversity, and its rich morphology, alongside the shortage of data resources \cite{shaalan2019challenges}. This gap is particularly evident in domain-specific applications such as the calls domain, where the need for high-accuracy ASR systems is critical for a range of applications from customer service to emergency response communications \cite{haraty2007casra+}.

Recent efforts in Arabic ASR have focused on addressing the challenges posed by the language's dialectical variations and the colloquial speech prevalent in inside and outside the telecommunication filed. Work in \cite{elnagar2021systematic} highlighted the significant impact of dialectal diversity on ASR performance, stressing the necessity for dialect-specific models and training datasets. Similarly, research by\cite{alsayadi2021arabic} underscored the importance of deep learning and neural network approaches in enhancing the accuracy of Arabic ASR systems, demonstrating promising advancements in handling the phonetic and morphological complexities of Arabic. While researchers in \cite{nasr2023end} attempted to enhance Arabic ASR on calls with the use of state-of-the-art ASR architectures.

Despite these technological advancements, the development of a comprehensive benchmark for Arabic ASR in the calls domain remains a critical need. Benchmarks play a pivotal role in evaluating and comparing the performance of ASR systems, yet the availability of such resources for Arabic, particularly for domain-specific applications, is limited \cite{kolobov2021mediaspeech}. The creation of a benchmark would not only facilitate the assessment of current ASR technologies but also drive future innovations by identifying gaps and areas for improvement as benchmarks serve as standardized tests that measure the performance of ASR systems under various condition \cite{alhanai2016development}.

The calls domain—encompassing telecommunications, customer service, and emergency response communications include handling low-quality audio, diverse accents and dialects, and the rapid, informal nature of spoken dialogue. While benchmarks exist for other aspects of Arabic ASR, focusing on general speech recognition or specific dialects \cite{mubarak2021qasr}, there remains a stark void when it comes to standardized testing environments that mimic the calls domain's intricate conditions.

This paper attempts to address this absence of benchmark to better evaluate available Arabic ASR systems, given the critical role that effective ASR technologies play in enhancing communication, accessibility, and automation within Arabic-speaking regions.

\section{Data Description and Preparation}

This section presents the process of data collection and preparation for the development of the benchmark:

\subsection{Dataset Description}

Collected from calls between agents and clients across the Arab region, our benchmark dataset is sourced from a Cloud Communication Suite Company which serves in the MENA region, with explicit user consent ensuring ethical compliance while also removing any sensitive information related to the business. This dataset encapsulates a rich fabric of the Arabic linguistic landscape, featuring six distinct Arabic dialects according to regions across various domains, including education, entertainment, and e-commerce, with a focus on positions such as customer support and sales. The speakers, poured from 13 different countries, bring a diverse demographic representation to the dataset, enriching it with a wide phonetic and morphological diversity inherent to their respective regions.

The audio quality of the dataset is standardized at a 16kHz sampling rate, mono channel, with 16-bit precision, catering to the technical requirements for clear speech recognition while accommodating a range of noise levels. These levels vary from clean to office clean, extending to noisy environments such as cars, with the distribution of these conditions evenly balanced to reflect real-world scenarios. This variability presents a comprehensive challenge for ASR systems, testing their adaptability across different acoustic settings.

The manual transcription process was meticulously carried out by a team of 41 annotators and further refined by 13 reviewers, culminating in a total of 132 hours of high-quality, annotated speech data. This rigorous annotation process ensures the dataset's utility in developing and evaluating ASR systems capable of navigating the complex linguistic and acoustic landscape of the Arab region. Through this benchmark, we aim to advance the field of Arabic speech recognition by providing a robust resource that mirrors the rich diversity and real-world conditions of Arabic speech communication.
The distribution of the regions in the benchmarks is as follows:

\begin{figure}
    \centering
    \includegraphics[width=0.5\textwidth]{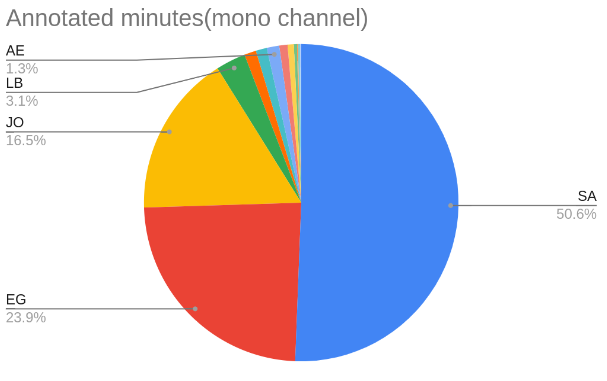}
    \caption{Distribution of the Regions}
    \label{fig:enter-label}
\end{figure}

\subsection{Data Preparation}
The following guidelines specifies the required methodology of annotation and verification, such as the exclusive use of Arabic characters, and the representation of numbers; whether to transcribe them as digits or spell them out in letters. This preparatory phase ensured consistency and accuracy in the annotated data, which is crucial for training and evaluating ASR systems effectively.

Following the establishment of annotation guidelines, audio segmentation was performed on the collected long-duration calls. This process involved dividing the calls into smaller, manageable chunks, each less than 30 seconds in length, based on periods of silence detected within the audio. This segmentation facilitated the annotation process, allowing for more focused and precise transcription of the audio content.

Post-annotation, the dataset underwent a rigorous quality assurance phase to ensure the integrity of both the audio and text components. This verification process included:

\begin{itemize} 

\item \textbf{Character Set Compliance:} A thorough review was conducted to ensure that all transcriptions adhered to the predefined character set requirements, maintaining consistency across the dataset.

\item \textbf{Audio File Integrity:} We systematically checked the dataset to eliminate extremely short or empty audio files, which hold minimal to no value for ASR training and evaluation.

\item \textbf{Automatic Language Detection:} To guarantee the linguistic purity of our dataset, each audio file was subjected to automatic language detection. This step was critical in confirming that all audios were in the Arabic language, aligning with our study's focus.

\item \textbf{Transcription Completeness:} Ensuring no audio files were left without a corresponding transcription was paramount. Each file was checked to confirm the presence of accurate and complete transcriptions, a crucial step for effective ASR evaluation.

\end{itemize}
Upon completing these checks, the dataset was stratified into levels based on the cleanliness and relevance to our specific domain. 

\section{Models}

In order to ensure that validity of the dataset and compare the performance of different ASR systems, 5 different state-of-the-art ASR systems were evaluated to establish a baseline for the developed benchmark.

\begin{itemize}

\item \textbf{Meta M4T Model}
Developed by Meta, M4T model \cite{barrault2023seamlessm4t} represents a significant advancement in multilingual ASR technology. Though specific details about its performance on Arabic speech are not widely published, the M4T model is designed to support multiple languages and dialects, leveraging large-scale datasets and deep learning to enhance its speech recognition capabilities. Its development reflects Meta's commitment to reducing language barriers and improving accessibility across its platforms.

The M4T models introduce a groundbreaking sentence embedding framework known as Sentence-level multi-modal and language-Agnostic representations, abbreviated as Sonar proposed by \cite{duquenne2023sonar} (see figure 2 below). This approach begins with the creation of a text embedding space, followed by the application of a teacher-student training methodology to incorporate the speech modality. Mirroring the Laser model's strategy \cite{artetxe2019massively}, Sonar's initial text space is built on an encoder-decoder framework, utilizing the NLLB-1.3B \cite{costa2022no} model which is adept at handling translations across 200 languages. The model modifies the intermediate representation to a fixed-size vector through mean-pooling, enabling the decoder to focus on a singular vector. This system is further refined with the comprehensive T2TT training dataset from NLLB, exploring various training objectives. An extensive analysis of this process reveals a robust, highly multilingual sentence representation capable of being decoded into any of the 200 languages covered by the NLLB initiative. This universal representation significantly enhances the performance of different NLP system especially speech recognition on low-resourced languages by leveraging embedding and pattern learnt from other higher-resourced languages.

\begin{figure}[!h]
    \centering
    \includegraphics[width=\textwidth, height=10cm]{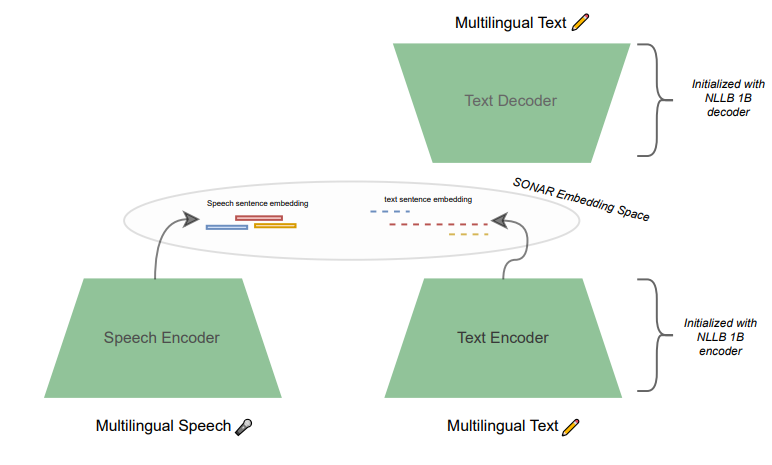}
    \caption{SONAR Architecture \cite{duquenne2023sonar}}
    \label{fig:enter-label}
\end{figure}

\item \textbf{Whisper}
Whisper \cite{radford2023robust} is an automatic speech recognition (ASR) system trained on 680,000 hours of multilingual and multitask supervised data collected from the web. OpenAI shows that the use of such a large and diverse dataset leads to improved robustness to accents, background noise and technical language compared to other architectures. Moreover, it enables transcription in multiple languages, as well as translation from those languages into English. OpenAI open-sourced models and inference code to serve as a foundation for building useful applications and for further research on robust speech processing.

The Whisper architecture is a simple end-to-end approach, implemented as an encoder-decoder Transformer. Input audio is split into 30-second chunks, converted into a log-Mel spectrogram, and then passed into an encoder. A decoder is trained to predict the corresponding text caption, intermixed with special tokens that direct the single model to perform tasks such as language identification, phrase-level timestamps, multilingual speech transcription, and to-English speech translation. Figure 3 depicts the blocks of Whisper ASR architecture.

\begin{figure}[!h]
    \centering
    \includegraphics[width=\textwidth, height=10cm]{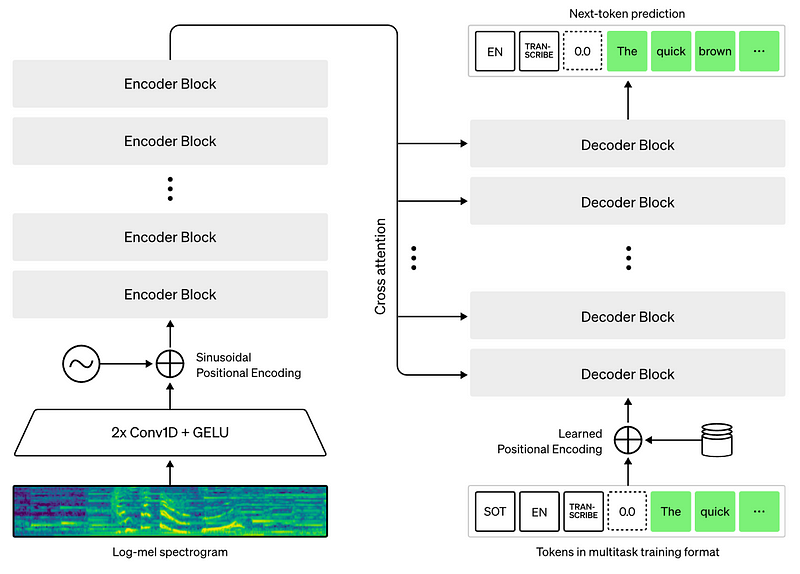}
    \caption{Whisper Encoder-Decoder Architecture \cite{radford2023robust}}
    \label{fig:enter-label}
\end{figure}

\item \textbf{Chirp}
is an ASR technology developed to specifically address the challenges of recognizing speech in noisy environments. While details about Chirp's underlying architecture and specific performance metrics are less publicly documented, it is part of Google's suite of speech technologies, known for their robustness and efficiency in processing speech from diverse sources and conditions.

\item \textbf{Google Cloud API \footnote{https://cloud.google.com/speech-to-text?hl=en}}
The Google Cloud Speech-to-Text API is a powerful tool that enables developers to convert audio to text by applying powerful neural network models. It supports a wide range of languages and dialects, including Arabic, and offers features such as real-time streaming transcription. Google's API is optimized for accuracy and can handle noisy audio files effectively, making it a popular choice for applications requiring high-quality speech recognition.

\item \textbf{Azure API\footnote{https://learn.microsoft.com/en-us/azure/ai-services/speech-service/rest-speech-to-text}}
Microsoft's Azure Speech to Text API is part of the Azure Cognitive Services, offering advanced speech recognition capabilities. This API supports multiple languages and dialects, including Arabic, with different dialects, customizable models, and robust handling of different audio environments. Azure's API is designed for scalability and integration into various applications, providing a reliable solution for developers needing speech-to-text services.

\end{itemize}

\section{Results and Discussion}

Results of the above models shown in the table below:

\begin{table}[!h]
\centering
\caption{Results of different ASR models}
\begin{tabular}{ccc}
\hline
\textbf{Model} & \textbf{WER} & \textbf{CER} \\ \hline
\textbf{Chirp}          & 48.9\%       & 22.4\%       \\
\textbf{Meta M4T V1}       & 67.8\%       & 34.3\%       \\
\textbf{Google API}     & 67.1\%       & 40.60\%      \\
\textbf{Azure API}      & 71.88\%      & 39.04\%      \\
\textbf{Whisper Large V1}        & 83.8\%       & 52.3\%       \\ \hline
\end{tabular}
\end{table}

The evaluation of Automatic Speech Recognition (ASR) systems reveals a wide range of performance across different models, as evidenced by their Word Error Rate (WER) and Character Error Rate (CER). These metrics serve as critical benchmarks for assessing the ability of ASR systems to accurately transcribe spoken language into text. Among the evaluated models, Chirp emerges as the clear leader, boasting the lowest WER at 48.9\% and CER at 22.4\%. This suggests that Chirp is significantly more effective at correctly identifying and transcribing words and characters, a capability likely bolstered by its design optimization for noisy environments. Such environments pose substantial challenges for ASR systems due to background noise and acoustic distortions, which can significantly impair the clarity of speech signals.

On the other end of the spectrum, Whisper recorded the highest WER and CER, at 83.8\% and 52.3\%, respectively. These figures indicate a considerable portion of words and characters were incorrectly transcribed by the system, highlighting its limitations, especially in complex auditory environments or perhaps with diverse accents and dialects.

Between these two extremes, the Meta M4T, Google API, and Azure API systems present a gradient of effectiveness. Meta M4T and Google API exhibited similar performance levels with Meta M4T at a WER of 67.8\% and CER of 34.3\%, closely followed by Google API at a WER of 67.1\% and a higher CER of 40.60\%. Azure API positioned itself slightly less favorably, with a WER of 71.88\% and CER of 39.04\%. These figures indicate that while these systems are capable of performing automatic speech recognition, their accuracy rates vary, affecting their reliability for applications requiring precise transcription.

The overall performance of these ASR systems underscores the complexity of speech recognition technology and the significant challenges it faces, particularly in adverse conditions. Factors such as background noise, speaker accents, dialect variations, and speech tempo all contribute to the difficulty of accurately transcribing spoken language. While Chirp's design allows it to excel in noisy settings, the overall findings highlight a crucial need for further research and development in this field. Enhancements in machine learning algorithms, audio processing, and linguistic modeling are essential to drive improvements in ASR technology, aiming to achieve higher accuracy rates and make these systems more adaptable to the vast variability of real-world speech.

\bibliographystyle{unsrt}  
\bibliography{references}

\end{document}